\documentclass[10pt]{article} 
\usepackage[preprint]{tmlr}

\usepackage{amsmath,amsfonts,bm}

\def\eqref#1{equation~\ref{#1}}

\def\1{\bm{1}}

\DeclareMathAlphabet{\mathsfit}{\encodingdefault}{\sfdefault}{m}{sl}
\SetMathAlphabet{\mathsfit}{bold}{\encodingdefault}{\sfdefault}{bx}{n}

\usepackage{hyperref}
\usepackage{url}

\usepackage{booktabs}
\usepackage{array}
\usepackage{graphicx}

\usepackage{tikz}
\usetikzlibrary{positioning,arrows.meta,fit,calc,shapes,decorations.pathreplacing}
\usepackage{subcaption}
\usepackage{array}
\usepackage{pifont}
\newcommand{\cmark}{\ding{51}} 
\newcommand{\xmark}{\ding{55}} 
\usepackage{multirow}
\usepackage{pdflscape}

\title{Open Problems in Differentiable Social Choice: Learning Mechanisms, Decisions, and Alignment}

\author{\name Zhiyu An \email zan7@ucmerced.edu \\
      \addr Department of Computer Science and Engineering\\
      University of California, Merced
      \AND
      \name Wan Du \email wdu3@ucmerced.edu  \\\addr Department of Computer Science and Engineering\\
      University of California, Merced
}

\def\month{02}  
\def\year{2026} 

\begin{document}

\maketitle

\begin{abstract}
Social choice has become a foundational component of modern machine learning systems. From auctions to resource allocation and the alignment of large generative models, machine learning pipelines increasingly aggregate heterogeneous preferences and incentives into collective decisions. In effect, many contemporary machine learning systems already implement social choice mechanisms, often implicitly and without explicit normative scrutiny.

Classical social choice and mechanism design offer deep axiomatic foundations for collective decision-making, but their traditional formulations are poorly matched to the scale, adaptivity, and distributional complexity of today’s socio-technical systems. In response, a rapidly growing literature has begun to reframe social choice, mechanism design, and alignment as differentiable learning problems, in which voting rules, allocation mechanisms, and incentive schemes are parameterized as neural or differentiable architectures, optimized from data, and evaluated empirically.

This Review synthesizes this emerging paradigm—what we term differentiable social choice—across auctions, decision aggregation, and AI alignment. We articulate a unifying perspective in which loss functions act as implicit aggregation rules, architectural inductive biases encode axioms such as anonymity or feasibility, and properties like robustness, proportionality, and incentive compatibility emerge as learned—but fragile—outcomes. 

Beyond synthesizing methods, this Review is explicitly forward-looking. Across three domains, we identify \textbf{18 concrete open problems} spanning incentive guarantees, robustness to distribution shift and strategic manipulation, certification and auditability of learned mechanisms, pluralistic preference aggregation, and the governance of alignment objectives. These open problems delineate a new research agenda at the intersection of machine learning, economics, and social choice theory.
\end{abstract}


\section{Introduction}

The intersection of artificial intelligence and social choice theory is undergoing a structural transformation. For much of the twentieth century, collective decision-making---elections, public good provision, auctions, and committee decisions---was treated as an axiomatic enterprise. Mechanisms were derived as closed-form objects from normative desiderata such as strategyproofness, Pareto efficiency, anonymity, or revenue optimality. This classical program yielded foundational insights and impossibility results, including Arrow’s theorem, Gibbard--Satterthwaite, and Myerson’s optimal auction theory \cite{arrow2020social,gibbard1973manipulation,satterthwaite1975strategy,myerson1981optimal}. At the same time, these results expose intrinsic limitations of the axiomatic approach. Many desirable properties are mutually incompatible, guarantees are typically worst-case and agnostic to the statistical distribution of real-world data, and the resulting mechanisms often rely on strong rationality and tractability assumptions that are violated in practice \cite{hart2013menu,caiata2025voting}.

Modern deployments have amplified these limitations. Spectrum auctions, online advertising markets, participatory budgeting, decentralized governance platforms, and preference-based training of large language models (LLMs) all operate in decision spaces that are high-dimensional, combinatorial, non-convex, and distribution-dependent.
Yet, social choice is (both implicitly and explicitly) used in many of the most important contemporary machine learning research, e.g. in automated auction mechanism design for revenue optimzation \cite{duan2023scalable}, in Large Language Models joint creative generations \cite{duetting2024mechanism}, and, most celebrated recently, in preference-based alignment (RLHF) of LLMs from human annotators who have inherently heterogeneous preference models \cite{siththaranjan2023dpl, swamy2024minimaximalist}.
We summarize these applications of social choice in machine learning in Table \ref{tab:intro-socialchoice-ml-ref}.
In such settings, analytic derivations with universal guarantees can be simultaneously too pessimistic---ruling out mechanisms that perform well on realistic data---and too optimistic---failing catastrophically under distribution shift or strategic adaptation \cite{hart2013menu,caiata2025voting}. Nevertheless, many contemporary applied systems are already learned: allocation policies, ranking rules, and reward models are trained by gradient descent, whether or not designers acknowledge their implicit social-choice content.

This tension has catalyzed the emergence of \emph{Differentiable Social Choice}. The central premise is that mechanisms need not be static artifacts of logic; instead, they can be parameterized, differentiable functions optimized from data, subject to explicit structural and normative constraints. Drawing on advances in differentiable economics \cite{dutting2017optimal,dutting2019optimal,anil2021learning}, neural social choice \cite{Matone2024DeepVoting,Hornischer2024}, and learning-based mechanism design \cite{fairstein2024learning,kesari2025fluid}, this paradigm treats voting rules, auctions, and allocation mechanisms as trainable objects. Classical axioms are no longer enforced solely by analytical proofs; they reappear as inductive biases, architectural symmetries, loss terms, and audit criteria.

Concrete applications already illustrate both the promise and the stakes of this shift. In auction design, neural mechanisms have been trained to approximate revenue-optimal auctions in multi-item settings where analytic solutions are unknown, encoding incentive compatibility as regret penalties or architectural constraints \cite{dutting2019optimal,ivanov2022optimal,hertrich2023mode}.  In AI alignment, preference-learning objectives used to train reward models and policies have been shown to implicitly instantiate specific voting rules---most notably Borda count---even when designers intend to remain agnostic about normative aggregation \cite{siththaranjan2023dpl,dai2024mapping}.

\begin{figure}
    \centering
    \includegraphics[width=\linewidth]{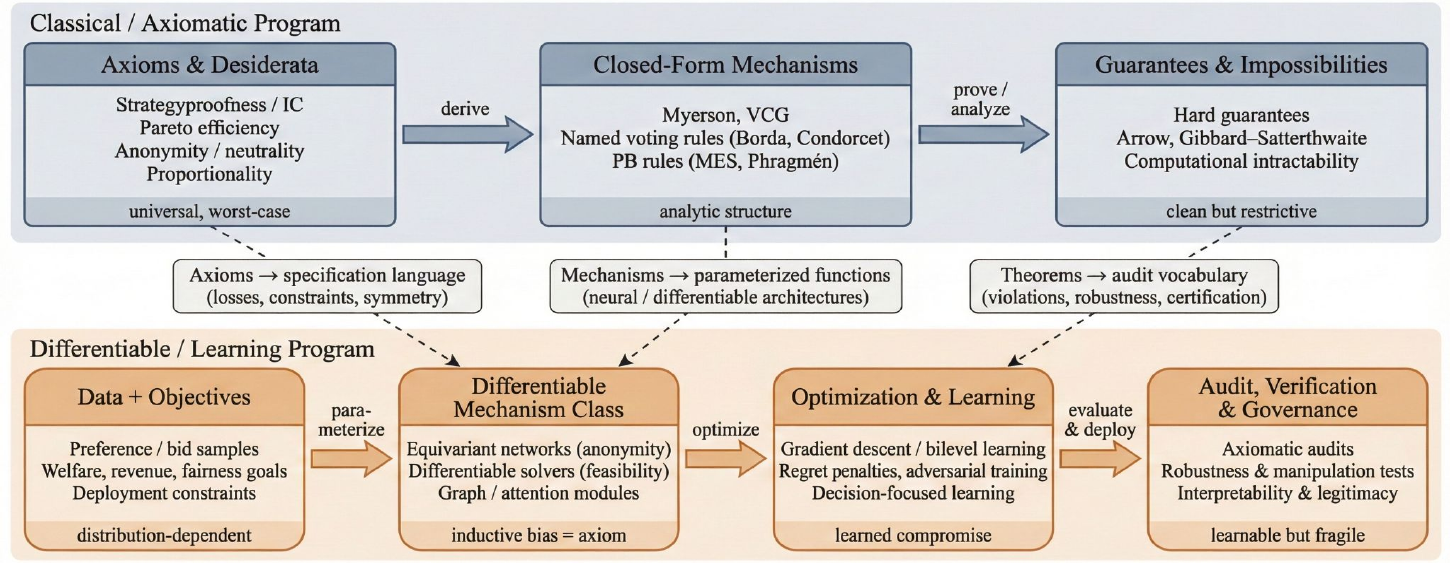}
    \caption{A structural shift from axiomatic social choice (closed-form mechanisms derived from normative axioms) to differentiable social choice (parameterized mechanisms optimized from data), where axioms reappear as architectural biases, loss terms, and audit criteria.}
    \label{fig:conceptual_shift}
\end{figure}

These developments mark a deep conceptual shift. We structurally illustrate this conceptual shift in Figure \ref{fig:conceptual_shift}. Social choice axioms become a \emph{specification language} for loss functions and architectures, and an \emph{audit vocabulary} for interpreting learned systems. Anonymity and neutrality translate into permutation-equivariant neural networks; feasibility and budget constraints become differentiable solver layers; incentive compatibility is relaxed into approximate regret bounds that can be empirically evaluated and stress-tested. At the same time, the differentiable turn exposes new risks and novel research gaps: soft constraints may fail silently, expressive models may hide violations, and optimization objectives may encode unintended normative commitments.

\begin{table}[t]
\centering
\scriptsize
\setlength{\tabcolsep}{6pt}
\renewcommand{\arraystretch}{1.2}
\caption{\textbf{Where social choice appears in modern ML systems.}
Many modern ML pipelines combine signals from multiple agents or sources in ways that are mathematically analogous to social choice or mechanism design principles. The “implicit aggregation rule” column points to how such aggregation arises in practice, and the final column gives representative references supporting these connections.}
\label{tab:intro-socialchoice-ml-ref}
\begin{tabular}{p{2.0cm} p{2.4cm} p{3.0cm} p{3.5cm} p{3.5cm}}
\toprule

\textbf{ML system} &
\textbf{Agents / sources} &
\textbf{Aggregated signal} &
\textbf{Implicit aggregation rule} &
\textbf{Representative reference} \\

\midrule

\textbf{Learned auction design} &
Strategic bidders &
Bids, valuations &
Optimization of revenue under incentive constraints creates an implicit choice over allocation and payment functions &
Automated auctions using neural networks and mechanism learning  \cite{duan2023scalable, dutting2021optimal} \\

\textbf{Neural voting / ranking models} &
Voters or users &
Rankings, approvals, clicks &
Learning to predict or rank outcomes aggregates individual preferences similar to positional and Condorcet rules &
\cite{chen2023breaking, boutilier2023modeling, suryanarayana2022explainability} \\

\textbf{Preference-based alignment (RLHF)} &
Human annotators &
Pairwise preference labels &
Aggregating pairwise comparisons resembles social choice rules such as Borda/Bra­dley-Terry mechanisms &
Mapping social choice theory to RLHF and preference aggregation \cite{siththaranjan2023dpl, dai2024mapping, conitzer2024social, swamy2024minimaximalist}\\
\bottomrule
\end{tabular}
\end{table}

This Review surveys this rapidly evolving landscape with a unifying perspective. We organize the literature into six interconnected domains: (1) \emph{Differentiable Economics}, which studies learning-based approximations to optimal auctions and contracts; (2) \emph{Neural Social Choice}, which synthesizes, analyzes, and audits voting rules using deep learning; and (3) \emph{AI Alignment as Social Choice}, which interprets preference learning and reinforcement learning from human feedback (RLHF) as implicit voting.

As collective decision-making becomes embedded in machine learning pipelines, the design of loss functions, architectures, and evaluation protocols becomes inseparable from social choice theory. Differentiable social choice thus emerges not only as a technical research program, but as a foundational framework for the safety, interpretability, and auditability of algorithmic decision-making systems.

\section{Differentiable Economics: From Optimal Auctions to Distributed Edge Markets}
\label{sec:diff-econ}

Mechanism design is often described as ``inverse game theory'': rather than predicting equilibria of a given game, the designer chooses the game form so that equilibrium behavior induces a desirable outcome. In the classical theory, this program is driven by structural characterizations and closed-form solutions---most prominently, Myerson's optimal auction for a single item with independent private values \cite{myerson1981optimal}. Yet, even in stylized multi-item settings, optimal mechanisms can require high-dimensional (and potentially exponentially complex) menus, intricate randomization, and discontinuous allocation rules \cite{hart2013menu, hartline2009simple}. This gap between elegant theory and messy, high-dimensional practice has made automated mechanism design an enduring ambition in economics and computer science \cite{conitzer2002complexity,sandholm2003automated}.

Differentiable economics reframes this ambition as a learning problem: parameterize the allocation and payment rules as differentiable function approximators (typically neural networks), optimize expected revenue or welfare over samples from a valuation distribution, and enforce incentive constraints via differentiable surrogates \cite{dutting2017optimal, dutting2019optimal}. The conceptual shift is not merely computational convenience. It changes the object of study from axioms $\rightarrow$ closed-form mechanisms, to \emph{objectives + constraints} $\rightarrow$ learned mechanisms whose structure is discovered (and sometimes audited) empirically. This section reviews the methodological core, the rapidly evolving architectural choices (from multilayer perceptrons to attention and graph-based inductive biases), the emerging theory for optimization and verification, and the expansion from textbook auctions to resource-constrained, distributed edge markets.

\subsection{From analytical optimal auctions to learned mechanisms}
\label{subsec:learned-mechanisms}

\paragraph{Classical foundations and why multi-dimensionality is hard.}
In single-parameter environments (e.g., one item, or matroid constraints with single-dimensional types), incentive compatibility admits strong structural characterizations; for example, monotone allocations and threshold payments yield dominant-strategy incentive compatibility (DSIC) \cite{myerson1981optimal}. In contrast, in multi-parameter environments (multiple heterogeneous items, bundles, or quality levels), the feasible set of DSIC mechanisms is governed by \emph{cyclic monotonicity} of the allocation rule, and payments correspond to (sub)gradients of a convex potential \cite{rochet1987necessary}. This geometry is powerful but rarely yields closed forms outside special cases. Theoretical work shows that optimal multi-item mechanisms can be highly complex (sometimes requiring large menus or delicate randomization), making the search space difficult to reason about analytically \cite{hart2013menu}.

\paragraph{Neural parameterizations: allocation and payment networks.}
The differentiable approach implements a mechanism as two coupled maps: an \emph{allocation network} $x_\theta(\cdot)$ producing (possibly fractional or randomized) allocations, and a \emph{payment network} $p_\theta(\cdot)$ producing transfers. The earliest influential line, often associated with the ``deep automated mechanism design'' paradigm, trains these networks end-to-end to maximize expected revenue under incentive constraints \cite{dutting2017optimal, dutting2019optimal}. In practice, the allocation network is often designed to satisfy feasibility by construction (e.g., softmax-based assignment for unit-demand, differentiable sorting/ranking for position auctions, or projection layers for budgets), while payments are produced either by a separate network or by integrating along value paths when exploiting Rochet-style structure \cite{dutting2017optimal,dutting2019optimal,rochet1987necessary}.

\paragraph{Recovering known solutions and discovering new ones.}
A key empirical validation is that learned mechanisms can recover canonical solutions where theory is complete (e.g., Myerson-like reserve pricing in single-item auctions) and outperform strong baselines in settings where the optimal mechanism is unknown \cite{dutting2017optimal, dutting2019optimal}. This is not simply curve-fitting: the learned mapping must simultaneously satisfy feasibility, approximate DSIC/BIC, and optimize revenue, which tends to expose where classical analytical templates (e.g., simple itemwise pricing) leave money on the table in correlated or multi-item settings \cite{dutting2019optimal,huo2022learning}.

\subsection{Differentiable incentive constraints: regret, cyclic monotonicity, and certification}
\label{subsec:ic}

\paragraph{Regret minimization as a differentiable surrogate.}
The dominant technical obstacle is that incentive constraints are universal quantifications over deviations. For DSIC, truthful reporting must maximize utility for every agent and every profile of others' bids. Regret-based training replaces this with a differentiable penalty:
\begin{equation}
\mathrm{Regret}_i(\theta; v) \;=\; \max_{\hat v_i} \; u_i\!\big(\hat v_i, v_{-i}; \theta \big) \;-\; u_i\!\big(v_i, v_{-i}; \theta \big),
\end{equation}
where $u_i(\hat v_i, v_{-i}; \theta)$ denotes agent $i$'s quasi-linear utility under a misreport $\hat v_i$ when the other agents report $v_{-i}$ and the mechanism is parameterized by $\theta$, while $u_i(v_i, v_{-i}; \theta)$ is the utility under truthful reporting; the regret $\mathrm{Regret}_i(\theta; v)$ therefore measures the maximum utility gain achievable by unilateral deviation from truthfulness at valuation profile $v$.
Regret-based training optimizes expected revenue subject to low expected regret, often via Lagrangian or augmented Lagrangian schemes \cite{dutting2017optimal}. This family of methods is frequently associated with RegretNet-style pipelines, in which an inner maximization (adversarial misreport search) is approximated by gradient ascent on $\hat v_i$, while an outer minimization updates mechanism parameters \cite{dutting2017optimal}.

A subtlety, highlighted by subsequent work, is that regret estimation itself can be unreliable: the inner maximization is non-convex and can under-estimate true regret, producing mechanisms that appear incentive compatible during training but violate constraints under stronger attacks \cite{you2026bridging}. This has motivated better adversarial optimizers, validation protocols tailored to regret-based mechanisms, and objective formulations that expose a more interpretable trade-off between revenue and acceptable IC violation (e.g., regret budgets) \cite{ivanov2022optimal,you2026bridging}.

\paragraph{RochetNet and structure from cyclic monotonicity.}
An alternative to pure regret penalization is to enforce DSIC via structural conditions derived from convex analysis. Rochet's theorem provides a necessary and sufficient condition for implementability in quasilinear multi-parameter domains: an allocation is DSIC implementable iff it is cyclically monotone, and payments can be derived from a convex potential \cite{rochet1987necessary}. RochetNet-style constructions exploit this by parameterizing a convex function and defining allocations/payments through its gradients, achieving \emph{architectural} incentive properties rather than purely penalty-based ones \cite{dutting2019optimal,hertrich2023mode}. In practice, such approaches trade expressivity and optimization simplicity against stronger built-in guarantees.

\paragraph{Certification and verification: beyond empirical audits.}
As differentiable mechanisms move toward deployment, the field has begun to explore \emph{certification}: given a trained network, can we prove it is DSIC (or approximately DSIC within a bound) over a domain? ``Certifying strategyproof auction networks'' develops tooling for post-training verification in restricted settings, reflecting a broader push to reconcile black-box learning with the normatively central requirement of incentive guarantees \cite{curry2020certifying}. Verification remains challenging because IC is a global property over a continuous type space and can be brittle under distribution shift or stronger deviation models than those used in training \cite{you2026bridging}.

Table~\ref{tab:ic-enforcement} summarizes the main approaches to enforcing incentive compatibility in differentiable mechanism design, highlighting how they differ in enforcement mechanism, strength of guarantees, computational cost, and characteristic failure modes.

\begin{table}[t]
\caption{\textbf{Incentive enforcement strategies in differentiable mechanism design.}
Methods differ in where incentive constraints enter (training-time penalties vs.\ architectural implementability vs.\ post-hoc certification), the strength of guarantees (empirical expected regret vs.\ structural/certified bounds), and characteristic failure modes (notably regret underestimation and distribution-shift brittleness).}
\label{tab:ic-enforcement}
\centering
\scriptsize
\setlength{\tabcolsep}{4pt}
\renewcommand{\arraystretch}{1.25}
\begin{tabular}{p{1.75cm} p{1.40cm} p{2.35cm} p{1.50cm} p{1.50cm} p{1.45cm} p{1.55cm} p{2.80cm}}
\toprule
\textbf{Approach} &
\textbf{IC target} &
\textbf{Enforcement mechanism} &
\textbf{Optimization structure} &
\textbf{Guarantee type} &
\textbf{Compute profile} &
\textbf{Expressivity} &
\textbf{Typical failure modes / best use} \\
\midrule

\textbf{Regret penalties} \cite{dutting2017optimal,dutting2019optimal} &
Approx.\ DSIC / BIC &
Add regret term $\mathbb{E}[\max_{\hat v_i} u_i(\hat v_i)-u_i(v_i)]$; inner-loop deviation search &
Bilevel min--max; Lagrangian / ALM &
Empirical / in-distribution (expected regret on samples) &
High (inner maximization per batch) &
High (generic NN mechanisms) &
\textbf{Failure:} regret underestimation when the inner maximizer is weak; IC can break under stronger attacks or shift \cite{you2026bridging}. \textbf{Best:} flexible baseline for high-dimensional auctions. \\

\textbf{Stronger deviation solvers} \cite{ivanov2022optimal,you2026bridging} &
Approx.\ DSIC / BIC &
Improve inner maximization (multi-start, longer ascent, better attacks); sometimes regret budgets &
Still bilevel; often tighter outer objectives &
Still empirical (but more reliable stress-tests) &
Very high (more attack steps / restarts) &
High &
\textbf{Failure:} remains attacker-class dependent; can overfit to chosen deviation model. \textbf{Best:} when IC violations are costly and compute is available. \\

\textbf{Architectural IC (cyclic monotonicity / convex potential)} \cite{rochet1987necessary,hertrich2023mode} &
DSIC implementability (domain-dependent) &
Parameterize convex potential; derive allocation/payment via gradients to satisfy implementability conditions &
Single-level training (no explicit inner max) &
Structural (architecture-level, within modeled domain) &
Moderate &
Medium (restricted class) &
\textbf{Failure:} mismatch to domain assumptions; limited mechanism family; feasibility layers can still be brittle. \textbf{Best:} when built-in IC is required and structure is acceptable. \\

\textbf{Structured mechanism templates} \cite{li2024deep} &
Template-dependent (often approximate IC) &
Constrain to interpretable families (e.g., rank-score rules, parametric payments) and learn parameters &
Single-level ERM / decision-focused &
Partial / template-inherited &
Low--moderate &
Low--medium &
\textbf{Failure:} may leave revenue/welfare on the table; hidden violations if template assumptions fail. \textbf{Best:} deployments needing interpretability, regulation fit, or low latency. \\

\textbf{Post-hoc certification / verification} \cite{curry2020certifying} &
DSIC or bounded approx.\ DSIC &
Verify trained network over a domain (e.g., partitions / discretization / bounds) &
Train-then-certify pipeline &
Certified bound on checked domain &
Potentially high (scales with dimension/coverage) &
Any (but certifiable only in restricted settings) &
\textbf{Failure:} scalability and coverage gaps; guarantees may not extend beyond tested region. \textbf{Best:} high-stakes auditing and compliance. \\

\textbf{Robustness-aware training (distributional / ambiguity)} \cite{ball2024robust} &
IC + performance under shift &
Optimize worst-case or neighborhood objectives; ambiguity sets / stability refinements &
Min--max or robust ERM variants &
Model-dependent robustness guarantees &
High--variable &
Medium--high &
\textbf{Failure:} misspecified ambiguity sets can induce performance cliffs; robustness can trade off sharply with revenue. \textbf{Best:} settings with known drift/uncertainty in bidder distributions. \\

\bottomrule
\end{tabular}
\end{table}

\subsection{Architectures and inductive biases: symmetry, sets, and attention}
\label{subsec:architectures}

\paragraph{Why inductive bias matters in economic learning.}
Mechanisms are typically \emph{anonymous} (invariant to permuting bidder identities) and often \emph{neutral} with respect to symmetric items. Generic multilayer perceptrons must learn these symmetries from data, wasting capacity and risking spurious identity leakage. This motivates architectures that enforce permutation equivariance/invariance by construction \cite{rahme2021permutation}.

\paragraph{Permutation-equivariant RegretNet variants.}
A prominent line introduces permutation-equivariant layers for auction networks, showing both sample-efficiency and revenue gains by baking symmetry into the model class \cite{rahme2021permutation}. ``Optimal-er Auctions through Attention'' advances this theme by replacing MLP blocks with attention modules (RegretFormer) and proposing a regret-budgeted loss that makes the revenue--IC trade-off more controllable \cite{ivanov2022optimal}. Attention is not only a modern deep learning convenience; it offers a principled way to represent auctions as \emph{set-to-set} mappings where interactions among bidders and items must be modeled flexibly while respecting exchangeability.

\paragraph{Transformers for combinatorial and joint auctions.}
Recent work pushes beyond bidder symmetry to the deeper challenge of combinatorial structure. In joint auctions (e.g., simultaneous allocation of multiple related opportunities), both cross-bidder and cross-item dependencies can be high-order and context dependent. Transformer-based designs such as JTransNet (and related CAFormer/CANet-style variants) use self-attention to model these interactions, naturally handling variable-sized sets and enabling anonymous deterministic allocation in joint auction settings \cite{zhang2025transformer}. Empirically, these approaches often dominate older baselines in revenue while improving scalability across varying numbers of bidders/items, at the cost of increased training complexity and the need for careful constraint handling \cite{zhang2025transformer}.

\paragraph{Domain-guided architectures in ad auctions and industrial marketplaces.}
A complementary industrial thread constrains the mechanism class to improve interpretability and deployment fit. For example, rank-score allocation rules and second-price-style payments can be parameterized and learned end-to-end to optimize platform metrics in e-commerce advertising, producing mechanisms that remain close to Generalized Second Price templates while benefiting from data-driven tuning \cite{li2024deep}. This ``structured learning'' view suggests a continuum between fully flexible neural mechanisms and tightly constrained parametric families---with the right point on the continuum determined by deployment needs (transparency, latency, regulation) as much as by revenue.

\subsection{Optimization landscapes and theory: generalization, connectivity, and expressivity}
\label{subsec:theory}

\paragraph{Generalization bounds and the role of sampling.}
Because objectives are expectations over valuation distributions, differentiable economics naturally invites statistical learning theory: how many samples are needed for near-optimal revenue, and how does approximation error scale with network capacity and constraint enforcement? Early deep auction design work provides formal generalization bounds under regret-based formulations, clarifying when learned mechanisms should transfer from training draws to unseen valuations \cite{dutting2017optimal,dutting2019optimal}. These results are only partial---they typically bound performance under the assumed distribution and do not, by themselves, guarantee robustness to misspecification.

\paragraph{Mode connectivity and benign non-convexity.}
A persistent concern is that training is non-convex and could converge to poor local optima. ``Mode Connectivity in Auction Design'' provides a theoretically grounded counterpoint: for RochetNet-like architectures (and generalized affine maximizer auctions), local minima are connected by simple low-loss paths in parameter space, implying that the effective landscape is surprisingly benign \cite{hertrich2023mode}. This connects differentiable mechanism design to a broader deep learning observation: many seemingly hard non-convex objectives exhibit connectivity among good solutions, which helps explain why stochastic gradient methods can reliably find high-quality mechanisms despite limited theoretical guarantees.

\paragraph{Expressivity meets economic complexity.}
From an economic perspective, benign optimization does not eliminate the deeper question: what class of mechanisms can a given architecture represent? Classical work shows that optimal mechanisms over general type spaces may be discontinuous, menu-complex, or rely on subtle randomization \cite{hart2013menu}. Contemporary theory on general type spaces further emphasizes that even defining the right structural template can be delicate when types are disconnected or non-convex \cite{prasad2025revenue}. Neural networks are attractive precisely because they can approximate complex mappings, but the tension remains between expressivity (capturing the true optimum) and verifiability (certifying IC, IR, fairness, and feasibility).

\subsection{Robustness and strategic adaptivity: from distribution shift to adversarial learning}
\label{subsec:robustness}

\paragraph{Distribution shift as a first-class economic concern.}
A learned mechanism is only as reliable as the distributional assumptions implicit in its training data. If bidder populations, correlations, or value supports change, revenue and incentive properties can degrade sharply. In economics, this motivates \emph{robust mechanism design}: optimize performance under ambiguity about the prior. ``Robust Robustness'' refines this perspective by observing that a mechanism designed for a worst-case ambiguity set can itself be fragile to misspecification of that set; it proposes a robustness notion that requires payoff guarantees to be stable in neighborhoods around the ambiguity set (in the weak topology), avoiding ``performance cliffs'' \cite{ball2024robust}. This refinement resonates strongly with modern machine learning concerns about out-of-distribution generalization and adversarially induced failures.

\paragraph{Strategic manipulation of the learning process.}
Beyond passive shift lies an active threat: agents may strategically manipulate not only their bids within a fixed mechanism, but also the seller's \emph{learning} of the mechanism over time. ``Adversarial Learning in Revenue-Maximizing Auctions'' studies repeated auctions where sellers adapt mechanisms based on past bids; bidders can learn bidding strategies that exploit this adaptivity even without full knowledge of the seller's update rule, improving bidder utility and potentially reducing seller revenue \cite{nedelec2019learning}. This perspective reframes differentiable economics as a \emph{learning-in-games} system: the seller runs gradient descent on mechanism parameters while bidders run their own learning dynamics, producing a coupled non-stationary environment. It also suggests an emerging security agenda: mechanism learning algorithms should be robust to strategic data poisoning by bidders, analogous to adversarial training in machine learning \cite{nedelec2019learning,you2026bridging}.

\paragraph{Correlated values and richer priors.}
Much of the early neural auction literature uses independent priors for tractability, but real markets commonly exhibit correlation (shared signals, common-value components, contextual effects). Learning-based approaches can incorporate such structure directly by sampling from correlated distributions and optimizing end-to-end, though incentive and identification issues can become more subtle \cite{huo2022learning}. This intersects with a broader trend in mechanism design: moving from analytically convenient priors to data-driven, high-dimensional priors inferred from platform logs.

\subsection{From auctions to edge markets: partial fulfillment and resource-constrained mechanism design}
\label{subsec:edge}

\paragraph{Why edge markets change the mechanism design problem.}
Edge computing markets allocate heterogeneous resources (CPU, GPU, memory, bandwidth, energy) across distributed service providers and application service providers. Unlike canonical item auctions, allocations can be \emph{divisible} and \emph{partially fulfilled}: an agent may receive 60\% of requested compute, with utility and feasibility governed by engineering constraints, latency, and network topology. Classical mechanisms such as VCG \cite{clarke1971multipart,vickrey1961counterspeculation,groves1973incentives} may be computationally burdensome or misaligned with partial fulfillment and operational constraints.

\paragraph{Neural mechanisms for partial fulfillment.}
Neural Auction Mechanisms for Partial Fulfillment in Edge Computing proposes architectures explicitly tailored to this domain, decoupling allocation decisions into probability and quantity components (PDNet) and using exchangeable attention mechanisms (EANet) to preserve bidder anonymity while scaling across variable numbers of agents \cite{zhang2026neural}. The learned mechanisms are trained with revenue objectives and regret-based incentive penalties, while embedding feasibility constraints relevant to edge capacity and service provisioning. A notable contribution is the modularity of the differentiable approach: by altering output parameterizations and constraint projections, the same learning framework can address allocation regimes that are awkward for traditional all-or-nothing auction templates \cite{zhang2026neural}.

\begin{figure}[t]
    \centering
    \includegraphics[width=\linewidth]{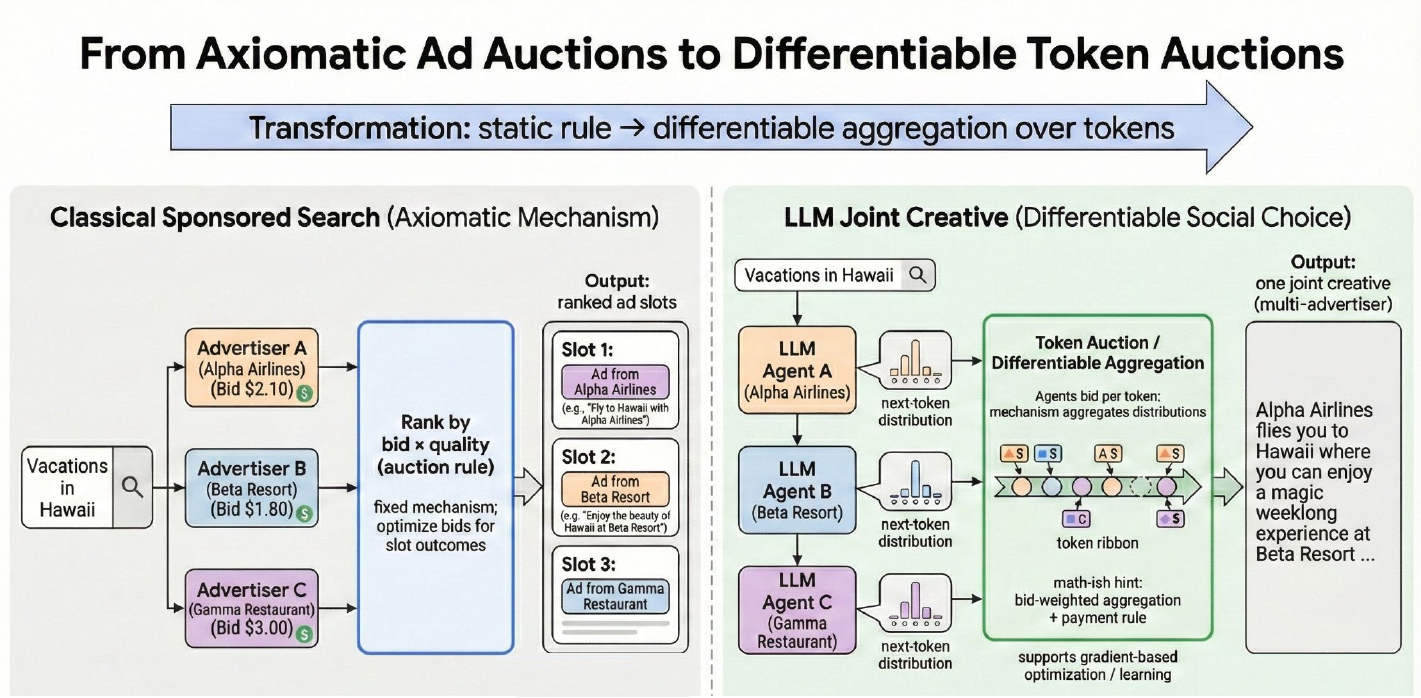}
    \caption{Comparison of Classical Ad Auction vs. LLM Joint Creative Mechanisms. On the left, a classical ad auction ranks advertisers by bid times quality to allocate fixed slots. On the right, LLM agents representing advertisers produce next-token distributions that are aggregated via a differentiable token auction mechanism. The aggregation proceeds token-by-token with bids influencing the combined output distribution; the resulting creative jointly reflects advertiser preferences and supports gradient-based learning.}
    \label{fig:auction_transformation_example}
\end{figure}

Complementing these neural mechanisms in edge settings, \textit{Mechanism Design for Large Language Models} \cite{duetting2024mechanism} proposes a token auction model for aggregating outputs from multiple self-interested LLM agents in joint content generation. Rather than allocating discrete items, this framework operates on a token-by-token basis, letting each LLM agent submit a single-dimensional bid that influences the aggregated next-token distribution in generation. We illustrate the process and compare it with classical Ad Auction process in Figure \ref{fig:auction_transformation_example}. \cite{duetting2024mechanism} formalize incentive properties for such mechanisms and show they are equivalent to a monotonicity condition on output aggregation, enabling analogues of second-price payments even without explicit valuation functions \cite{duetting2024mechanism}. They further derive concrete aggregation rules — e.g., bid-weighted averages in linear and log-space — that align joint generation outcomes with agents’ implicit preferences over distributions. This token auction approach not only bridges classical auction design with generative AI but also provides a differentiable mechanism that can be integrated with learning-based policies, illustrating how economic mechanisms and gradient-based optimization can jointly shape multi-agent LLM outputs.

\paragraph{Distributed implementation and the path to ``edge-native'' mechanism design.}
Edge settings also raise systems questions often absent in classical auction theory: decentralized information, limited communication, privacy constraints, and real-time decision requirements. These pressures motivate mechanisms that are lightweight to evaluate, stable under asynchronous updates, and robust to missing or delayed signals. While current work largely trains centralized networks and deploys them as decision policies, the architectural trend toward permutation-equivariant and set-based models is compatible with more distributed realizations (e.g., message passing or hierarchical attention), hinting at a convergence between differentiable mechanism design and distributed learning.

\subsection{Open Problems}
\label{subsec:diff-econ-open-problems}

\paragraph{\textit{OP1: }Hard incentive guarantees in expressive networks.}
How can we move from soft regret penalties and empirical audits to \emph{hard}, architecture-level DSIC (or certified approximate DSIC) guarantees for high-capacity models without collapsing expressivity?

\paragraph{\textit{OP2: }Reliable regret estimation under strong deviations.}
Can we design inner-loop deviation solvers and validation protocols that provably upper bound regret (rather than under-estimate it) in realistic continuous domains, especially for transformer-based mechanisms?

\paragraph{\textit{OP3: }Robustness to distribution shift and ambiguity misspecification.}
What is the right unification of distributionally robust mechanism design (including refinements such as robust robustness) with modern out-of-distribution learning so that learned mechanisms degrade gracefully under shifting bidder populations?

\paragraph{\textit{OP4: }Mechanism learning under strategic data poisoning.}
How do we design seller-side learning dynamics that are provably robust when bidders strategically manipulate bids to influence the \emph{future} mechanism, rather than only optimizing within a fixed mechanism?

\paragraph{\textit{OP5: }Edge-native decentralization and communication constraints.}
Can differentiable mechanisms be trained and deployed in a decentralized manner that respects edge communication limits and privacy constraints while maintaining feasibility and incentive properties?

\paragraph{\textit{OP6: }Normative constraints beyond revenue.}
How should we represent and enforce normative desiderata---fairness, diversity, market access, or sustainability---in differentiable objectives so that these constraints are not merely post-hoc metrics but are satisfied reliably at deployment time?


\section{Neural Social Choice and Voting Rules}
\label{sec:neural-social-choice}

Social choice theory studies how to aggregate individual preferences into a collective decision.
In its classical form, the subject is defined by axioms (normative desiderata) and impossibility
results: Arrow's theorem shows that no rank-based social welfare function can simultaneously
satisfy a natural set of fairness and rationality axioms once there are at least three
alternatives \cite{arrow2020social,sen1986social}. In the same spirit, Gibbard--Satterthwaite
establishes that every onto, non-dictatorial deterministic voting rule with at least three
alternatives is manipulable \cite{gibbard1973manipulation,satterthwaite1975strategy}.
These foundational barriers gave rise to a taxonomy of voting rules---plurality, Borda, Condorcet
extensions, approval, STV, and many others---each embodying an explicit compromise between
incompatible ideals \cite{black1958theory,Fishburn1974,Tideman2006}.

Neural social choice (and, more broadly, differentiable social choice) asks a different design
question: rather than selecting among a small menu of named rules, can we \emph{learn} an
aggregation rule from data, while steering it toward an intended axiomatic compromise?
This move is motivated by three pressures. First, many attractive rules are computationally
hard to compute exactly (e.g., Dodgson, Young, Kemeny--Young), making them difficult to
deploy at scale \cite{Bartholdi1989,Conitzer2007}. Second, worst-case
axiomatic analysis can be overly pessimistic: rule failures may be rare or common depending on
the distributional structure of preferences (e.g., single-peakedness, Mallows noise, or real-world
correlations) \cite{Marden1995,Procaccia2010,Elkind2016}. Third, modern
collective decisions increasingly happen in algorithmic settings (recommenders, ranking, RLHF,
and platform governance) where the ``electorate'' and its preferences are observed as data streams
rather than as a one-off election \cite{Gleich2011,Mattei2013}.

This chapter reviews the emerging literature around (i) \emph{representation and architectural
inductive bias} for preference profiles, (ii) \emph{mimicry} of classical rules versus \emph{synthesis}
of new rules, (iii) \emph{axiomatic auditing} and the generalization gap between outcome imitation
and principled behavior, (iv) \emph{distribution-specific} evaluation of rule violations in practice,
(v) robustness and adversarial training for manipulation resilience, and (vi) ``loss-as-rule''
frameworks that make the aggregation principle explicit and differentiable.

\subsection{Axioms, impossibilities, and the learning objective}
\label{subsec:axioms-impossibilities}

\paragraph{From binary axiom satisfaction to differentiable proxies.}
Classical voting axioms are typically discrete properties of a function $f$ mapping a preference
profile to an outcome (winner, ranking, or committee). Core examples include anonymity (voter
symmetry), neutrality (candidate symmetry), Pareto efficiency, monotonicity, Condorcet
consistency, and (for multi-winner rules) proportionality and representation guarantees
\cite{Moulin1988,Young1995,brandt2016handbook}. In neural settings, these
axioms are often evaluated empirically as \emph{violation rates} over sampled profiles, or encoded
as differentiable penalties that approximate discontinuous desiderata \cite{Hornischer2024,an2026differential}.

A central conceptual shift is that the object being designed is no longer a closed-form rule with
a known proof portfolio, but a function class (e.g., a permutation-invariant network) optimized
against a loss. This reframes axioms as (i) constraints to be satisfied (hard or soft), (ii) terms in
a multi-objective optimization, or (iii) evaluation metrics that can sharply diverge from the training
objective. The resulting design space is closer to ML model selection than to classical axiomatic
deduction, raising questions about transparency, validation, and normative governance.

\paragraph{Manipulation as a learning-and-security problem.}
Gibbard--Satterthwaite is not nullified by deep learning: any sufficiently expressive deterministic
rule that is non-dictatorial and onto remains manipulable in principle \cite{gibbard1973manipulation,satterthwaite1975strategy}.
What changes is the empirical and operational framing. In learned rules, we can:
(i) measure \emph{attack surfaces} and \emph{attack success rates} under bounded coalitions or
bounded vote changes, (ii) train with adversarial examples to reduce these rates, and (iii) trade
off manipulation resistance against other desiderata (e.g., welfare) via explicit objectives
\cite{Li2026Learning,Procaccia2010}.

\subsection{Representing preference profiles: symmetry, sets, and canonical embeddings}
\label{subsec:representations}

\paragraph{Why representation is the first mechanism.}
A preference profile for $n$ voters and $m$ candidates can be encoded as (i) a matrix of ranks,
(ii) pairwise comparisons, (iii) positional score vectors, or (iv) a multiset of permutations.
In learning, the encoding implicitly imposes structure: a naive one-hot tensor with fixed voter
indices risks breaking anonymity; similarly, candidate indexing can leak spurious biases if neutrality
is not respected. This motivates permutation-invariant/equivariant architectures as a form of
\emph{architectural axiom enforcement} \cite{Zaheer2017,Lee2019}.

\paragraph{Permutation-invariant neural social choice.}
Early work already observed a surprising phenomenon: standard neural classifiers trained to
predict winners often behave like simple positional rules (notably Borda), even when trained on
other targets, suggesting strong inductive biases induced by common encodings and optimization
dynamics \cite{Burka2016,Armstrong2021}. ``Learning to Elect'' systematizes this by
casting voting rules as set-input functions and comparing DeepSets, Set Transformers, and graph
networks for (a) mimicking classical rules and (b) discovering welfare-maximizing rules
\cite{anil2021learning}. The key takeaway is that set-invariant models offer both theoretical
alignment with anonymity/neutrality and practical gains in sample efficiency and generalization.

\paragraph{Canonical embeddings from social choice theory (DeepVoting).}
DeepVoting pushes representation further by importing \emph{canonical embeddings} from social
choice into the learning pipeline \cite{Matone2024DeepVoting}. Instead of learning directly from raw
rank matrices, DeepVoting uses embeddings derived from positional scoring vectors and related
structures, and recasts deterministic rules as \emph{probabilistic} social choice functions that output
distributions over candidates. This probabilistic relaxation is not merely an ML trick: it smooths
discontinuities, supports gradient-based fine-tuning for normative properties, and naturally aligns
with settings where collective decisions are inherently stochastic (e.g., randomized tie-breaking,
sampling-based committees, or probabilistic recommendation). A central empirical claim is that
tailored embeddings reduce model size requirements and improve scaling to larger electorates
relative to prior black-box encodings \cite{Matone2024DeepVoting}.

\subsection{Mimicry versus synthesis: what does it mean to ``learn a voting rule''?}
\label{subsec:mimicry-synthesis}

\paragraph{Mimicking hard-to-compute rules.}
One pragmatic motivation for neural social choice is computational: approximate a rule that is
normatively appealing but costly to compute (e.g., Kemeny--Young, Dodgson, Young) by a fast
forward pass \cite{Bartholdi1989,Conitzer2007}. ``Learning to Elect''
demonstrates that modern set-invariant networks can mimic both positional (plurality, Borda) and
more complex rules at high accuracy, with generalization depending critically on distribution shift
and profile size changes \cite{anil2021learning}. DeepVoting similarly targets efficient learning
of complex rules through structured embeddings \cite{Matone2024DeepVoting}. This line is valuable
as an \emph{approximation infrastructure}---but it also exposes a conceptual pitfall: matching outputs
on typical training distributions does not imply the learned function satisfies the axioms that
\emph{define} the target rule.

\paragraph{Synthesis: learning rules that optimize welfare or axioms directly.}
A second, more radical agenda is to treat the rule itself as a decision policy optimized for a
population and a criterion. ``Learning to Elect'' trains networks to maximize utilitarian or
egalitarian welfare under assumed voter utility models, and reports learned rules that outperform
classical ones on the target welfare measure, matching known optima in special cases
\cite{anil2021learning}. DeepVoting extends this idea through \emph{axiomatic fine-tuning}:
a network pretrained to mimic a standard rule can be updated (via gradient descent) to reduce
pathologies such as a probabilistic version of the no-show paradox, effectively ``patching'' a rule's
weaknesses while keeping much of its behavior intact \cite{Matone2024DeepVoting,Moulin1988NoShow}.
This illustrates a distinctive promise of differentiable design: rather than choosing between Borda
and Condorcet families, one can navigate a continuous space of hybrid rules between them.

\subsection{Axiomatic auditing and the ``mimicry--reasoning'' gap}
\label{subsec:audit-gap}

\paragraph{Axiomatic Deep Voting (Learning How to Vote with Principles).}
Hornischer and Terzopoulou provide a critical ``reality check'' by developing an axiomatic
audit framework for neural voting rules \cite{Hornischer2024}. They report a
disconnect between (i) high predictive accuracy in mimicking a target rule's outcomes and
(ii) low satisfaction of that rule's defining axioms under distribution shift or out-of-distribution
tests. Notably, they argue that training on axiom-specific data does not automatically repair
axiom satisfaction. Instead, they show that \emph{directly optimizing} axiom satisfaction (through
axiom-derived loss terms) can synthesize novel neural rules that occupy new regions of the
axiom trade-off landscape and can substantially differ from any named classical rule
\cite{Hornischer2024}.

\paragraph{Why the gap appears: discontinuity, rare constraints, and compositionality.}
Several mechanisms plausibly contribute to the mimicry--reasoning gap. Many voting rules are
piecewise-constant with complex decision boundaries; small perturbations near boundaries can
flip winners, making generalization fragile in high-dimensional profile spaces. Moreover, axioms
often constrain behavior on \emph{rare} configurations (e.g., specific paradoxical profiles), so
finite-sample mimicry may never ``see'' the regions where axioms bite. Finally, axioms are
global and compositional (they quantify over counterfactual profiles), while supervised training
is pointwise. The audit-first perspective thus becomes essential: learned social choice cannot be
assumed to inherit normative properties from their labels \cite{Hornischer2024}.

\paragraph{Interpretability and explanation of learned rules.}
A complementary response is to \emph{explain} learned rules in human-legible terms. ``Learning to
Explain Voting Rules'' learns interpretable surrogates (e.g., trees) that approximate voting behavior
and can expose which profile features drive outcomes, supporting governance and debugging
\cite{Kang2023}. This connects neural social choice to the broader ML agenda of
post-hoc interpretability, but with unusually crisp normative stakes: explanations are not only for
trust but for verifying that the rule embodies the intended compromise.

\subsection{What voting rules actually do in practice: distribution-specific social choice}
\label{subsec:data-driven-axioms}

\paragraph{From worst-case theorems to empirical violation rates.}
Much of social choice theory is worst-case: an axiom is either satisfied for all profiles or not.
Caiata, Armstrong, and Larson argue that this binary view can be misleading for deployed systems
that operate under structured preference distributions \cite{caiata2025voting}. They
propose a data-driven evaluation framework for multi-winner voting rules that estimates how
frequently axioms are violated across synthetic and real-world distributions. Their analysis suggests
that classical rules with clean theoretical stories may violate desirable properties frequently on
realistic distributions, while learned neural rules can be trained to reduce these violation rates for
a target population \cite{caiata2025voting}.

\subsection{Robustness and manipulation resilience: adversarial learning for elections}
\label{subsec:robustness-elections}

\paragraph{Adversarial GNNs for resilient elections.}
Li, Shah, and Giusti explicitly connect learned voting to AI security by proposing adversarially
trained graph neural networks for elections \cite{Li2026Learning}. They represent an election
as a bipartite graph (voters--candidates) and learn the aggregation function with a GNN, which
naturally encodes anonymity and neutrality via permutation invariance over nodes. They introduce
a min--max training regime in which an attacker perturbs votes (e.g., via strategic manipulation
or noise injection) to flip outcomes, and the model is trained to maintain target properties such as
welfare and robustness under these attacks \cite{Li2026Learning}.

\paragraph{Why graph structure is a natural inductive bias.}
Graph representations unify several desiderata: (i) exchangeability (node permutation), (ii)
variable-size generalization to different numbers of voters/candidates, and (iii) a clear interface for
modeling coalitions or local perturbations as graph edits. This architecture choice aligns with
earlier set-invariant approaches \cite{Zaheer2017,Lee2019,anil2021learning}
but emphasizes \emph{attack modeling} and robustness objectives. The broader research question
is whether robustness metrics in elections should resemble adversarial robustness in classifiers,
or whether domain-specific constraints (e.g., bounded coalition size, bounded Kendall distance)
provide a more normatively meaningful threat model \cite{Procaccia2010}.

\subsection{Differential Voting: loss functions as voting rules}
\label{subsec:differential-voting}

\paragraph{Loss geometry encodes normative aggregation.}
A striking bridge between neural social choice and ML practice is the observation that many
learning pipelines \emph{implicitly} implement voting rules. In preference-based learning (including
RLHF reward modeling), the standard Bradley--Terry--Luce (BTL) log-likelihood corresponds to
a particular aggregation principle closely related to Borda-like scoring \cite{an2026differential}.
Differential Voting makes this connection explicit by constructing differentiable, instance-wise
losses whose population optima correspond to distinct classical rules, and by analyzing how
loss geometry (margin sensitivity, boundary concentration) translates into axiomatic behavior
\cite{an2026differential}.

\paragraph{Differentiable surrogates for Condorcet-style aggregation.}
Differential Voting proposes smooth surrogates for discontinuous criteria such as Copeland
(pairwise majority wins) and Kemeny (minimizing pairwise disagreement), and establishes
consistency and limiting behavior as smoothing vanishes \cite{an2026differential}. Conceptually,
this turns ``choose a voting rule'' into ``choose a loss'' and enables \emph{mix-and-match} social choice:
weighted sums of losses can interpolate between ideals (e.g., Condorcet consistency vs.\ smooth
optimization stability). This approach also clarifies a central methodological tension in neural
mechanism design: objectives that are normatively attractive often induce non-smooth or
combinatorial optimization, while smooth objectives may bake in unintended normative biases.

\subsection{Open Problems}
\label{subsec:neural-social-choice-open}

\paragraph{\textit{OP7: }Certification of axioms for learned voting rules.}
How can we provide post-training certificates (or architecture-level guarantees) that a learned
rule satisfies anonymity, neutrality, monotonicity, or Condorcet consistency beyond finite-sample
testing?

\paragraph{\textit{OP8: }Generalization under electorate shift.}
What training objectives and validation protocols ensure that axiomatic behavior persists when the
preference distribution changes (new candidates, new voter blocs, or different cultural contexts)?

\paragraph{\textit{OP9: }Normative specification and governance.}
Who chooses the axiom mix, loss weights, and robustness threat model when the rule is learned,
and how can these choices be made transparent and contestable?

\paragraph{\textit{OP10: }Robustness to strategic coalitions with realistic constraints.}
Can adversarial training be grounded in social-choice-native threat models (coalition size, limited
information, bounded rationality) rather than generic perturbation norms?

\paragraph{\textit{OP11: }Interpretable structure discovery.}
Can we reliably extract human-legible descriptions (e.g., scoring vectors, pairwise tournaments,
committee scoring rules) from trained networks so that the learned rule can be debated and
audited like a classical rule?

\paragraph{\textit{OP12: }Bridging single-winner, multi-winner, and continuous outcomes.}
Is there a unified differentiable framework that scales from single-winner elections to committees,
rankings, and continuous allocations while preserving the relevant axioms in each regime?

\section{AI Alignment as a Social Choice Problem}
\label{sec:alignment-social-choice}

\begin{figure}[t]
    \centering
    \includegraphics[width=\linewidth]{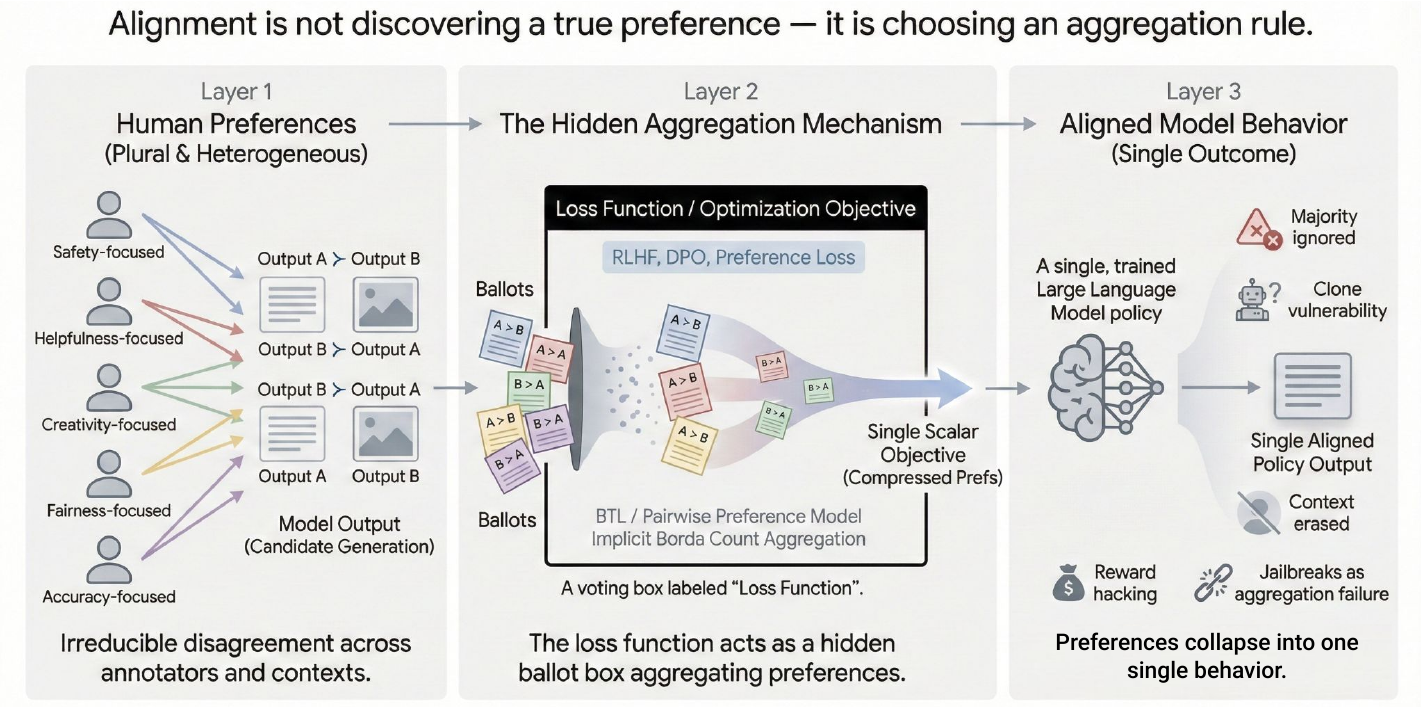}
    \caption{\textbf{AI alignment as a social choice problem.} Human annotators provide heterogeneous and often conflicting pairwise judgments over candidate model outputs (left), which are aggregated by alignment objectives such as RLHF or DPO through a preference-based loss function that acts as a hidden ballot box (center), implicitly corresponding to a particular social choice rule (e.g., BTL-induced Borda aggregation); optimization then collapses this plural input into a single trained policy and output (right), importing classical social-choice pathologies—such as majority insensitivity, context erasure, and vulnerability to manipulation—into alignment pipelines.}
    \label{fig:alignment_as_social_choice}
\end{figure}

\subsection{Section overview: the hidden ballot in the loss function}
\label{subsec:hidden-ballot}

Modern alignment pipelines for large language models (LLMs) are, at their core, \emph{preference aggregation systems}. Reinforcement Learning from Human Feedback (RLHF) learns from many pairwise comparisons---``output $A$ is better than output $B$''---collected from annotators who may disagree for principled, contextual, or idiosyncratic reasons \cite{christiano2017deep,stiennon2020summarize,ouyang2022instructgpt}. Direct Preference Optimization (DPO) and related ``direct alignment'' objectives bypass explicit reward modeling but still optimize a loss induced by preference data \cite{rafailov2023dpo}. In both cases, the training procedure compresses heterogeneous human judgments into a single reward model or policy, thereby performing an implicit social choice operation.
Figure \ref{fig:alignment_as_social_choice} illustrates how modern alignment objectives implement an implicit social choice rule, compressing heterogeneous human preferences into a single optimized behavior via the loss function rather than discovering a uniquely “correct” reward.

Recent theoretical work makes this connection explicit and consequential. Distributional Preference Learning (DPL) proves that standard Bradley--Terry--Luce (BTL) preference modeling---a workhorse in RLHF---implicitly aggregates across heterogeneous annotator contexts via the \emph{Borda count} \cite{siththaranjan2023dpl}. This equivalence is not merely rhetorical: it imports classical social choice pathologies (e.g., indifference-to-majority phenomena and clone vulnerabilities) into alignment, and it reframes ``reward hacking'' and ``jailbreaks'' as failures of aggregation under hidden context \cite{siththaranjan2023dpl}. In parallel, Conitzer et al.\ argue that social choice theory should guide alignment precisely because disagreement is irreducible: the problem is not to find a single ``true'' preference, but to choose a principled way to aggregate diverse feedback into system behavior \cite{conitzer2024social}. Dai and Fleisig sharpen the mapping by emphasizing where RLHF differs from canonical voting settings (e.g., strategic feedback collection, non-stationary policies, and distribution shift), cautioning against naive transfer of theorems while endorsing social-choice-native desiderata as analytical tools \cite{dai2024mapping}.

This chapter reviews alignment as a social choice problem across five themes:
(i) the social choice structure of RLHF and direct alignment objectives,
(ii) heterogeneity-aware learning (distributional and pluralistic alignment),
(iii) axiomatic and welfare-theoretic perspectives on what ``collective preference'' should mean,
(iv) strategic behavior, manipulation, and mechanism design for feedback pipelines, and
(v) governance and decentralization: from ``one reward model'' to constitutional and participatory alignment.

\subsection{RLHF as implicit voting: BTL, Borda, and the geometry of preference learning}
\label{subsec:rlhf-borda}

\paragraph{From pairwise labels to a social ranking.}
RLHF typically fits a reward model $r_\theta(x,y)$ so that, for a prompt $x$ and two candidate outputs $y^+$ and $y^-$,
\begin{equation}
\Pr(y^+ \succ y^- \mid x) \approx \sigma\!\big(r_\theta(x,y^+) - r_\theta(x,y^-)\big),
\end{equation}
which is the BTL likelihood used widely in preference learning \cite{christiano2017deep,stiennon2020summarize}. When labels come from many annotators and contexts, the learning objective averages log-likelihood contributions across the dataset; the critical question is: \emph{what aggregation principle does this averaging implement}?

DPL answers this precisely. Modeling ``hidden context'' as a latent variable indexing annotator type, rubric, or situational factors, Siththaranjan et al.\ show that minimizing the standard BTL negative log-likelihood implicitly aggregates these latent contexts according to Borda count \cite{siththaranjan2023dpl}. In classical social choice terms, Borda assigns each alternative a score equal to its average rank, favoring consensus candidates and smoothing over sharp disagreement \cite{brandl2019axiomatic,saari1994geometry}. In alignment terms, this predicts a familiar empirical outcome: models that feel like ``the safe average'' can arise not (only) from explicit safety tuning, but from the statistical structure of the loss.

\paragraph{Alignment objectives as social choice rules.}
A central message of this section is that modern alignment pipelines are not value-neutral optimization procedures: every preference-learning or direct-alignment objective implicitly defines a collective decision rule over candidate model behaviors.
Table~\ref{tab:alignment_as_social_choice} makes this connection explicit by mapping commonly used alignment losses to their corresponding social choice interpretations and axiomatic properties.
In particular, widely used RLHF-style objectives based on Bradley--Terry--Luce likelihoods implement Borda-like aggregation under heterogeneous or hidden contexts, inheriting known violations of majority responsiveness and Condorcet consistency.
Pluralistic approaches modify the structure of the electorate but do not, by themselves, resolve aggregation pathologies.
By contrast, Differential Voting specifies the aggregation principle directly at the level of the loss, enabling differentiable surrogates for Condorcet-consistent rules and making normative trade-offs explicit and auditable.

\paragraph{Social choice pathologies become alignment failure modes.}
The Borda lens imports a mature theory of failure. Borda can select an outcome that is not a Condorcet winner (an option that beats every other option head-to-head) \cite{condorcet1785}. It is susceptible to clone effects, where adding similar options can change the winner, and it can underweight intense majority preference in favor of broad second-choice acceptability \cite{Tideman2006}. Translating to RLHF, one expects:
(i) \emph{majority dissatisfaction} when a large subgroup strongly prefers a style or value that is diluted by averaging,
(ii) \emph{minority erasure} when rare but important preferences are smoothed away, and
(iii) \emph{strategic vulnerabilities} because if contributors understand the aggregation rule, they can shape outcomes by how they label and which comparisons they provide \cite{siththaranjan2023dpl}.

Dai and Fleisig emphasize an additional geometric mismatch: in voting, candidates are discrete; in RLHF, outputs lie on a vast, structured manifold and the model changes as data is collected \cite{dai2024mapping}. This makes classical impossibilities (e.g., Arrow-style) suggestive but not directly dispositive; nevertheless, they remain valuable as warnings that no aggregation rule will satisfy all desiderata simultaneously \cite{arrow2020social,sen1986social}.

\subsection{From ``one reward'' to pluralism: distributional and heterogeneous preference learning}
\label{subsec:pluralism}

\paragraph{Distributional Preference Learning (DPL).}
DPL proposes to represent preferences not by a single scalar score per output but by a distribution over scores, thereby preserving uncertainty and heterogeneity that BTL-style averaging collapses \cite{siththaranjan2023dpl}. Operationally, DPL introduces estimators that can detect when hidden context materially affects comparisons (e.g., via explained-variance diagnostics), and it demonstrates that capturing such variation can reduce downstream vulnerabilities such as jailbreak-like behaviors that exploit ``regression to the mean'' \cite{siththaranjan2023dpl}. The core conceptual contribution is that \emph{heterogeneity is not noise to be averaged away}; it is part of the target specification.

\paragraph{Pluralistic Alignment Framework (PAL) and ideal-point models.}
PAL advances pluralism from a diagnostic to an architectural principle. Rather than fitting a universal preference model, PAL uses latent-variable structure (ideal-point models and mixture modeling) to learn a shared preference embedding space together with personalized or subgroup-conditioned reward functions \cite{chen2024pal}. This connects alignment directly to political science and psychometrics, where ideal-point models have long been used to represent heterogeneous preferences with low-dimensional latent structure. PAL shows that this approach can be efficient (leveraging penultimate-layer representations) and can few-shot generalize to new users, which is a natural analogue of ``representation learning for electorates'' \cite{chen2024pal}.

\paragraph{Welfare interpretation: average-case optimality versus worst-case regret.}
Heterogeneity-aware alignment raises a normative question: if preferences differ, what does it mean to align to ``the people''? Shirali et al.\ formalize alignment with \emph{user types} and analyze the limits of learning a single policy under heterogeneous rewards \cite{shirali2025direct}. They show conditions under which the best single policy corresponds to maximizing average reward over types---a utilitarian aggregation---but also highlight that achieving this may require information that is often missing in practice (e.g., annotator identity or type labels) \cite{shirali2025direct}. This tension mirrors a classical divide in social choice between utilitarian welfare aggregation and protections against minority harm \cite{harsanyi1955cardinal,rawls1971theory}.

\subsection{Direct alignment objectives as tournament solutions: DPO, multi-turn extensions, and unobserved heterogeneity}
\label{subsec:dpo-tournaments}

\paragraph{DPO as ``implicit reward modeling.''}
DPO shows that, under a KL-regularized RLHF objective, one can optimize the policy directly using preference pairs without explicitly fitting a reward model \cite{rafailov2023dpo}. Social-choice-theoretically, DPO can be viewed as selecting a policy that performs well in an induced tournament of outputs, where each preference comparison supplies a directed edge. This makes DPO a natural arena for importing tournament solution concepts (Condorcet, Copeland, maximal lotteries) and their differentiable surrogates \cite{brandt2016handbook}.

\paragraph{Unobserved preference heterogeneity and the need for richer feedback.}
Chidambaram et al.\ analyze DPO under \emph{unobserved} heterogeneity and argue that pairwise comparisons alone can be insufficient: when different subpopulations have systematically different preferences, binary labels can confound signal in ways that cannot be disentangled without additional structure, motivating \emph{ternary} or otherwise enriched feedback \cite{chidambaram2025direct}. This aligns with a broader identification theme: social choice outcomes depend not only on the aggregation rule but on what the ballot elicits. In alignment, ``ballot design'' includes the comparison protocol, annotator instructions, and the set of candidate outputs shown.

\paragraph{Mixture models and EM-style direct alignment.}
A complementary approach is to explicitly posit latent preference types and learn a mixture of policies or a conditioned policy---conceptually moving from a single-winner election to a multi-winner or personalized system. This agenda is prominent in recent DPO extensions that use EM-style procedures to jointly infer latent types and optimize type-specific policies, providing a bridge between pluralistic alignment and direct policy optimization \cite{chidambaram2025direct,shirali2025direct}. Such methods can be interpreted as learning a \emph{committee} of aligned behaviors, with a subsequent choice rule (e.g., user-conditional selection) deciding which behavior to deploy.

\paragraph{Multi-turn and segment-level alignment as structured aggregation.}
In interactive systems, preferences apply to trajectories, not isolated responses. SDPO proposes segment-level credit assignment for multi-turn dialogues, focusing optimization on key segments rather than individual turns (too local) or whole sessions (too coarse) \cite{kong2025sdpo}. This is naturally understood as a social choice design problem over structured objects: the ``candidates'' are dialogue trajectories, and the aggregator must respect temporal structure and long-horizon coherence.

\subsection{Axiomatic alignment: making the aggregation principle explicit}
\label{subsec:axiomatic-alignment}

\paragraph{From accidental Borda to designed aggregation rules.}
The key lesson of the Borda equivalence is not that Borda is ``bad,'' but that \emph{alignment already implements a voting rule}---often unintentionally \cite{siththaranjan2023dpl}. The natural response is axiomatic: specify which properties we want the aggregation to satisfy and design losses or training pipelines that implement them \cite{conitzer2024social}. In the alignment context, desiderata commonly include:
\begin{itemize}
    \item \textbf{Majority responsiveness} (respecting strong majority preferences),
    \item \textbf{Minority protection} (avoiding systematic erasure),
    \item \textbf{Monotonicity and consistency} under additional feedback,
    \item \textbf{Robustness} to manipulation and distribution shift,
    \item \textbf{Transparency} of what trade-off is being optimized.
\end{itemize}
Dai and Fleisig caution that translating axioms requires care: RLHF is an online, non-stationary learning system with endogenous candidate generation, so classical axioms may need adapted definitions \cite{dai2024mapping}.

\paragraph{Differential Voting and loss-design as rule specification.}
Differential Voting provides a constructive methodology: define the aggregation rule via a differentiable loss that approximates a desired voting criterion (e.g., soft Copeland or soft Kemeny), enabling gradient-based optimization toward Condorcet-like behavior without discrete search \cite{an2026differential}. Although introduced in the voting-rule setting, the same principle applies to alignment: if preference learning and direct alignment objectives are tournament aggregators, then loss design is mechanism design. This viewpoint also connects to the rapidly expanding ecosystem of ``direct alignment'' objectives (e.g., KTO and related human-aware losses), which reinterpret alignment losses through behavioral utility models \cite{ethayarajh2024kto}.

\paragraph{Constitutional AI as principle aggregation.}
Constitutional AI replaces many low-level preference labels about harmfulness with a smaller set of high-level principles (a ``constitution''), using model-generated critiques and revisions guided by these principles \cite{bai2022constitutional}. Social-choice-theoretically, this changes the ballot: from pairwise comparisons to normative constraints. It can be interpreted as moving from preference aggregation to \emph{judgment aggregation} over principles and their instantiations, where consistency and completeness become central concerns \cite{list2002judgment}. The open challenge is that a constitution is itself a social choice object: whose principles, which conflicts, and what tie-breakers?

\subsection{Strategic feedback, poisoning, and mechanism design for alignment pipelines}
\label{subsec:strategic-alignment}

\paragraph{Annotators are strategic agents once they understand the aggregation rule.}
A distinctive aspect of alignment is that the ``electorate'' (annotators, red-teamers, users) may have incentives to steer the model. DPL formalizes this by showing that under hidden context and implicit Borda aggregation, annotators can have incentives to misreport preferences to influence the learned reward function \cite{siththaranjan2023dpl}. This reframes data collection as mechanism design: the platform must choose incentives, sampling strategies, and aggregation rules that remain robust when participants attempt to manipulate outcomes.

\paragraph{Content moderation and strategic adaptation as adjacent alignment mechanisms.}
Strategic responses are not limited to alignment datasets. Content moderation systems induce an explicit game between a filter and users who adapt content to pass the filter. Ahmadi et al.\ analyze this as a mechanism design problem balancing freedom of speech against social distortion (manipulation cost), showing NP-hardness of optimal trade-offs and providing approximation and generalization guarantees from finite data \cite{ahmadi2025strategic}. The connection to alignment is methodological: both are instances of learning under strategic adaptation, where the deployed system changes the distribution of future inputs and feedback.

\paragraph{Robust aggregation and pluralistic defenses.}
Robustness can be pursued by (i) adopting aggregation rules that are less manipulable than Borda-like averaging (e.g., Condorcet-consistent or minimax-regret approaches), (ii) explicitly modeling preference heterogeneity so that attackers cannot exploit the ``mean'' behavior, and (iii) designing feedback protocols that elicit richer information (ternary or contextual labels) to reduce identification failures under heterogeneity \cite{chidambaram2025direct,siththaranjan2023dpl}. These strategies parallel robust statistics: defend the estimator (aggregation) by changing both the objective and the data-collection process.

\subsection{Open Problems}
\label{subsec:alignment-open-problems}

\paragraph{\textit{OP13: }From empirical audits to certified axiomatic guarantees.}
Can we provide post-training certificates that an alignment procedure implements a chosen aggregation principle (e.g., Condorcet-consistency analogues) under realistic assumptions about candidate generation and distribution shift?

\paragraph{\textit{OP14: }Minority protections without personalization collapse.}
How can we design pluralistic alignment methods that protect minority values without devolving into unstable over-personalization or enabling targeted manipulation of subgroup models?

\paragraph{\textit{OP15: }Ballot design for alignment.}
What feedback primitives (pairwise, ternary, contextual rationales, principle-based critiques) are information-theoretically sufficient to identify a socially desirable aggregation outcome under heterogeneous preferences?

\paragraph{\textit{OP16: }Strategic resistance in data collection and deployment.}
How should incentives, sampling policies, and aggregation rules be co-designed so that users and annotators cannot strategically steer models toward harmful or partisan behaviors?

\paragraph{\textit{OP17: }Constitution selection as a meta-social-choice problem.}
If constitutional alignment depends on a set of principles, how should those principles be chosen, updated, and legitimately governed when societies disagree?

\paragraph{\textit{OP18: }Bridging welfare objectives and safety constraints.}
Can we reconcile welfare-theoretic aggregation (utilitarian averages, max-min fairness, regret minimization) with hard safety constraints in a way that is transparent about trade-offs and robust to misspecification?

\begin{landscape}

\begin{table}[t]
\centering
\scriptsize
\setlength{\tabcolsep}{4pt}
\begin{tabular}{p{3.0cm} p{7.0cm} p{2.3cm} p{2.5cm} c c c c c}
\toprule
\multirow{2}{*}{Method} &
\multirow{2}{*}{Loss / Objective} &
\multirow{2}{*}{Setting} &
\multirow{2}{*}{Corresp. Voting Rule} &
\multicolumn{4}{c}{Axioms / Guarantees} &
\multirow{2}{*}{Diff.} \\
\cmidrule(lr){5-8}
& & & &
\begin{tabular}[c]{@{}c@{}}Majority\\ resp.\end{tabular} &
\begin{tabular}[c]{@{}c@{}}Pareto\\ (PO)\end{tabular} &
IIA/LIIA &
\begin{tabular}[c]{@{}c@{}}Condorcet\\ (PMC)\end{tabular} &
\\
\midrule

\multicolumn{9}{l}{\textbf{(A) Standard alignment objectives as implicit voting rules}}\\
\midrule

\textbf{RLHF reward model (BTL / logistic)} &
$\sum_{(x,y^+,y^-)} \log\!\Big(1+\exp\big(r_\theta(x,y^-)-r_\theta(x,y^+)\big)\Big)$ &
Pairwise preferences (RLHF) &
Borda-like scoring &
\xmark & \cmark & \xmark & \xmark & \cmark \\

\textbf{Exponential pairwise loss} &
$\sum_{(x,y^+,y^-)} \exp\big(r_\theta(x,y^-)-r_\theta(x,y^+)\big)$ &
Pairwise preferences (RLHF-style) &
Borda-like scoring &
\xmark & \cmark & \xmark & \xmark & \cmark \\

\textbf{Hinge / margin ranking loss} &
$\sum_{(x,y^+,y^-)} \max\!\Big(0,\,1+r_\theta(x,y^-)-r_\theta(x,y^+)\Big)$ &
Pairwise preferences (ranking / RLHF-style) &
Borda-like scoring &
\xmark & \cmark & \xmark & \xmark & \cmark \\

\textbf{Direct Preference Optimization (DPO)} &
$\mathbb{E}_{(x,y^+,y^-)}\!\Big[-\log\sigma\!\big(\beta(\log\pi_\theta(y^+|x)-\log\pi_\theta(y^-|x))\big)\Big]$
&
Direct policy optimization &
Tournament solution (loss-induced) &
\emph{var.} & \emph{var.} & \xmark & \emph{var.} & \cmark \\

\midrule
\multicolumn{9}{l}{\textbf{(B) Heterogeneity-aware / pluralistic alignment (changes the ``electorate'')}}\\
\midrule

\textbf{Distributional Preference Learning (DPL)} &
Latent-context model; preserves heterogeneity in preference signal (beyond a single scalar reward) &
Hidden-context preferences &
Borda-like aggregation in the BTL baseline &
\xmark & \cmark & \xmark & \xmark & \cmark \\

\textbf{Pluralistic Alignment Framework (PAL)} &
Latent-variable / ideal-point style preference model; user- or subgroup-conditioned reward/policy &
Mixture of preference types &
Mixture/committee-style aggregation &
\emph{type-dep.} & \emph{type-dep.} & \xmark & \emph{type-dep.} & \cmark \\

\midrule
\multicolumn{9}{l}{\textbf{(C) Axiomatic loss design (Differential Voting)}}\\
\midrule

\textbf{Soft Copeland} &
$-y\,s_{\tau,\beta}(\Delta)+\tfrac{\lambda}{2}\Delta^2$ &
Axiomatic loss for pairwise aggregation &
Copeland (smooth surrogate) &
\cmark & \cmark & \xmark & \cmark & \cmark \\

\textbf{Soft Kemeny} &
$\sigma(-y\,\Delta/\tau)$ &
Axiomatic loss for pairwise aggregation &
Kemeny (smooth Kendall surrogate) &
\xmark & \cmark & \xmark & \cmark & \cmark \\

\textbf{Majority-based objectives (family)} &
Any objective depending only on pairwise majority directions &
Pairwise-majority tournament &
(Pairwise) Majority aggregation &
\cmark & \cmark & \xmark & \cmark & \xmark \\

\textbf{LCPO (leximax Copeland subj.\ to PO)} &
Leximax Copeland with Pareto constraint (implemented via LP feasibility checks) &
Hard-constraint rule layer &
Copeland (PO-constrained leximax) &
\cmark & \cmark & \xmark & \cmark & \xmark \\

\bottomrule
\end{tabular}
\caption{\textbf{Alignment objectives as implicit social choice rules.}
The choice of loss in RLHF-style preference learning (and related direct alignment objectives) instantiates a concrete aggregation principle with specific axiomatic trade-offs.
BTL-style objectives correspond to Borda-like scoring under heterogeneous/hidden contexts \cite{siththaranjan2023dpl}, inheriting known violations of majority responsiveness and Condorcet consistency \cite{brandt2016handbook,ge2024axioms}.
DPO can be interpreted as optimizing a loss-induced tournament solution over sampled candidates, so axiom satisfaction is generally sampling- and protocol-dependent \cite{rafailov2023dpo,dai2024mapping,conitzer2024social}.
Differential Voting constructs differentiable surrogates for Condorcet-style rules (e.g., Copeland, Kemeny), making the intended aggregation principle explicit at the loss level \cite{an2026differential}.}
\label{tab:alignment_as_social_choice}
\end{table}
\end{landscape}

\bibliography{main}
\bibliographystyle{tmlr}

\end{document}